\documentclass[lettersize,journal]{IEEEtran}
\usepackage{amsmath,amsfonts}
\usepackage{algorithmic}
\usepackage{adjustbox}
\usepackage{algorithm}
\usepackage{array}
\usepackage[caption=false,font=normalsize,labelfont=sf,textfont=sf]{subfig}
\usepackage{textcomp}
\usepackage{stfloats}
\usepackage{url}
\usepackage{verbatim}
\usepackage{graphicx}
\usepackage{cite}
\begin{document}

\title{Advancements in Road Lane Mapping: Comparative Fine-Tuning Analysis of Deep Learning-based Semantic Segmentation Methods Using Aerial Imagery}

\author{Xuanchen (Willow) Liu, Shuxin Qiao, Kyle Gao, Hongjie He, Michael A. Chapman, Linlin Xu, and Jonathan Li
\thanks{Manuscript received April 19, 2021; revised August 16, 2021.}}

\markboth{Journal of \LaTeX\ Class Files,~Vol.~14, No.~8, August~2021}%
{Shell \MakeLowercase{\textit{et al.}}: A Sample Article Using IEEEtran.cls for IEEE Journals}


\maketitle

\begin{abstract}
This research addresses the need for high-definition (HD) maps for autonomous vehicles (AVs), focusing on road lane information derived from aerial imagery. While Earth observation data offers valuable resources for map creation, specialized models for road lane extraction are still underdeveloped in remote sensing. In this study, we perform an extensive comparison of twelve foundational deep learning-based semantic segmentation models for road lane marking extraction from high-definition remote sensing images, assessing their performance under transfer learning with partially labeled datasets. These models were fine-tuned on the partially labeled Waterloo Urban Scene dataset, and pre-trained on the SkyScapes dataset, simulating a likely scenario of real-life model deployment under partial labeling. We observed and assessed the fine-tuning performance and overall performance. Models showed significant performance improvements after fine-tuning, with mean IoU scores ranging from 33.56\% to 76.11\%, and recall ranging from 66.0\% to 98.96\%. Transformer-based models outperformed convolutional neural networks, emphasizing the importance of model pre-training and fine-tuning in enhancing HD map development for AV navigation.
\end{abstract}

\begin{IEEEkeywords}
Road lane extraction, aerial imagery, deep learning, semantic segmentation, transfer learning.
\end{IEEEkeywords}



\section{Introduction}

The rapid development of autonomous driving technologies highlights the critical role of high-definition (HD) maps, essential for the safe and efficient operation of autonomous vehicles (AVs). These maps provide precise environmental details, such as traffic signals, road features, and lane markings, enabling AVs to navigate accurately. While Earth observation data, like LiDAR and satellite imagery, are valuable for HD map creation, they have limitations such as uneven point distribution and lower spatial resolution. Aerial imagery from drones or UAVs offers a superior alternative with higher resolution and broader coverage, capturing lane markings more effectively.

Integrating AI and deep learning is crucial for processing this data and automating the extraction of key features, although challenges remain in dense urban environments. Traditional remote sensing methods for lane marking extraction are limited, whereas computer vision advances in semantic segmentation offer promising but underexplored solutions for this task.

This study addresses the gap by conducting a comparative analysis of twelve deep learning-based semantic segmentation models tailored to road lane extraction from aerial imagery. Using transfer learning on a partially labeled dataset, the research evaluates the models' performance across two datasets, providing a benchmark for future work in HD map development for AVs.
\section{Related Works}

Road lanes are essential components of road infrastructure, delineating paths for vehicle movement and facilitating smooth traffic flow while conveying important traffic regulations. These lanes are marked with various symbols on the road surface, which can differ significantly in shape, size, length, and color, reflecting the diversity of traffic rules and cultural norms globally. From dashed lines to solid lines, and from arrows to pedestrian crossings, each marking serves a specific purpose and is geometrically designed with clear boundaries to ensure visibility.

\subsection{Datasets for Lane Segmentation}
In the context of enhancing lane segmentation model performance, selecting appropriate training datasets is crucial. These datasets must provide high-resolution images that reveal detailed lane features, offer annotations for various lane types, be readily accessible for research purposes, and be sufficiently large to support effective model training and validation.

HD map data is mainly collected using mobile mapping systems equipped with sensors like LiDAR. Ground-level datasets such as the TuSimple Lane Detection Challenge Dataset \cite{yoo_2020}, the Road and Lane Dataset from the Karlsruhe Institute of Technology and Toyota Technological Institute (KITTI) \cite{geiger_2013}, the California Institute of Technology (Caltech) Lanes Dataset \cite{Chao_2019}, and the Berkeley Deep Drive 100K (BDD100K) dataset \cite{Yu_2020} are common but lack comprehensive coverage for extensive traffic management, often hindered by limited field of view and physical obstructions, making urban mapping laborious and resource-intensive. In contrast, aerial image datasets are crucial for large-scale traffic management systems. Despite various available sources, detailed annotations for lane markings are scarce. Notable datasets like the International Society for Photogrammetry and Remote Sensing (ISPRS) Potsdam and Vaihingen dataset \cite{Rottensteiner_2014} provide high-resolution images but lack lane-specific information, while Massachusetts Roads \cite{Azimi_2019} and SpaceNet datasets \cite{van_etten2021} offer urban annotations without sufficient detail for precise segmentation.

The SkyScapes dataset stands out with its 13 cm resolution images and detailed annotations across 12 classes of lane markings in both urban and suburban environments. It is freely accessible, making it an invaluable resource for accurate detection and classification of road lanes from an aerial perspective.

\subsection{Existing Semantic Segmentation Models}
In computer vision, semantic segmentation is crucial for complex real-world applications, offering pixel-wise labeling that divides images into segments corresponding to different objects or regions. Deep learning has transformed this field by moving from traditional clustering and contour-based methods to more precise \cite{Weinland_2011}\cite{Sonka_2013}, pixel-level segmentation, thereby enhancing accuracy and scene understanding \cite{Guo_2018}.

Semantic segmentation models are typically divided into Convolutional Neural Network (CNN)-based and Transformer-based. Significant advancements in CNNs, particularly through the development of Encoder-Decoder architecture, have greatly influenced semantic segmentation \cite{Badrinarayanan_2017}\cite{Ronneberger_2015}. Models like the Fully Convolutional Network (FCN) modify CNNs to handle images of any size by substituting fully connected layers with convolutional layers for size-invariant segmentation \cite{Long_2015}. Other notable models include FastFCN, which introduces Joint Pyramid Upsampling to merge multi-scale features efficiently \cite{Wu_2019}; U-Net, known for its skip connections and symmetric encoder-decoder structure which preserve spatial context \cite{Ronneberger_2015}; MobileNetV3, which optimizes for mobile devices by integrating Hardware-aware Network Architecture Search (NAS) and the NetAdapt algorithm \cite{Howard_2019}; ANN (Asymmetric Non-local Neural Networks), which reduces computational demands through pyramid sampling \cite{Zhu_2019}; and DeepLabV3 with its atrous convolutions for enhanced receptive fields \cite{Chen_2017}. Additional advancements include DeepLabV3+, which adds an encoder-decoder structure for refined edge detailing \cite{Chu_2021}, PSPNet that incorporates a Pyramid Pooling Module for aggregating multi-scale contextual information \cite{Zhao_2017}, and SegNeXt, which introduces a convolutional attention mechanism to improve computational efficiency while maintaining segmentation accuracy \cite{Guo_2022}.

The landscape of semantic segmentation underwent further evolution with the introduction of Transformer-based models, which apply the self-attention mechanism to image data \cite{Dosovitskiy_2020} \cite{Ranftl_2021} \cite{Zheng_2021}, excelling at modeling long-range dependencies and capturing global context \cite{Strudel_2021}. Vision Transformer (ViT) treats images as sequences of patches, applying self-attention to capture complex spatial relationships \cite{Dosovitskiy_2020}. Innovations like the Twins-SVT streamline spatial attention mechanisms \cite{Chu_2021}, while models like SegFormer \cite{Xie_2021} and Swin Transformer \cite{Liu_2021} integrate advanced features that optimize segmentation tasks by efficiently processing images at varying resolutions and combining multi-scale features.

\subsection{Evolution of Lane Marking Detection Methods }
Deep learning has significantly advanced the analysis of aerial imagery for road extraction, a fundamental step towards accurate lane detection. Techniques such as the Multi-Feature Pyramid Network (MFPN) proposed by have effectively managed varying road widths \cite{Gao_2018} and incorporated semantic segmentation and tensor voting to connect fragmented road segments \cite{Gao_2019}. Furthermore, Generative Adversarial Networks (GANs) have facilitated the extraction of detailed road features, including pavement and centerlines, demonstrating deep learning’s expanding role in comprehensive road network mapping \cite{Zhang_2019}. 

Research on road lane detection using aerial imagery remains relatively sparse, yet Azimi et al. (2019) have made notable contributions with their innovations such as Aerial LaneNet \cite{Azimi_2018} and SkyScapesNet \cite{Azimi_2019}. Aerial LaneNet employs Symmetric Fully Convolutional Neural Networks (FCNN) enhanced with Wavelet Transforms, focusing primarily on the binary classification of lane markings against the background . Building upon this foundation, SkyScapesNet, which is based on the Fully Convolutional DenseNet (FC-DenseNet) architecture \cite{2017fcdense}, leverages the SkyScapes dataset to combine dense semantic segmentation with semantic edge detection, enabling the segmentation of multiple road lane classes and the identification of small-scale urban features. 

\section{Description of Training Datasets}

\subsection{SkyScapes Dataset}
The SkyScapes Dataset consists of a comprehensive collection of aerial images captured over Munich, Germany \cite{Azimi_2019}. This dataset provides a detailed aerial view of the city's extensive transportation infrastructure within its urban and rural areas. Captured with a helicopter-mounted Digital Single-Lens Reflex (DSLR) camera system, the SkyScapes Dataset includes 16 RGB images, each with a resolution of 5,616×3,744 pixels, offering a ground sampling distance of 13 cm per pixel. It spans an area of 5.7 km².

This dataset is manually annotated with 31 semantic categories, focusing on elements typical in urban settings, such as various types of roads, parking spaces, bikeways, sidewalks, buildings, and different vehicle types. It includes a detailed categorization of 12 different lane marking types, such as dash-lines, crosswalks, stop-lines, and parking zones.

\subsection{Waterloo Urban Scene Dataset}
Waterloo, located in Ontario, Canada, is part of the Waterloo-Kitchener area. It features an extensive network of roads and significant highways that connect it with the Greater Toronto Area and nearby regions. 

Derived from the readily available Waterloo Building Dataset, the Waterloo Urban Scene Dataset offers high-resolution aerial ortho imagery of the Waterloo region \cite{he_2022}. Covering an extensive area of 205.8 km² and providing a fine spatial resolution of 12 cm per pixel, this dataset is perfectly suited for urban and traffic semantic segmentation projects. 

To tailor the Waterloo Urban Scene Dataset for this research, we enhanced it with manually added annotations to establish a comprehensive ground truth, essential for assessing the developed model’s effectiveness across diverse datasets. Originally lacking specific traffic classifications, 14 new classes were introduced, including various road and lane markings, vehicles, sidewalks, crosswalks, and traffic islands. 

The annotation process was meticulously organized into three main categories to improve clarity and specificity: facility types (e.g., roads, sidewalks, traffic islands), road lane markings, and vehicles. This categorization reflects actual conditions and aids in precise model evaluation. An essential aspect of our annotation approach was the establishment of a priority system for overlapping classes in the imagery, crucial for resolving ambiguities where a pixel belongs to multiple classes. It is noteworthy that even if background elements are obscured by foreground components—for instance, when a vehicle overlays a single solid line—the background element, such as the line, is still labeled entirely with overlapping class polygons. 

\section{Methodology for Automated Lane Marking Extraction}
\subsection{Description of the General Workflow}
Both datasets underwent data augmentation and parameter calculation before entering the training phase. Original images were flipped and cropped for data augmentation. For parameter calculation, the mean and standard deviation were computed on the training images to normalize the inputs for the both training and validation phase, which helps convergence.

Following this, twelve comparative models were trained using the Skyscapes dataset and evaluated to produce performance metrics on the Skyscapes test dataset. Then, these twelve models, initialized with weights from their training on the Skyscapes dataset, were fine-tuned on the Waterloo Urban Scenes dataset, after which their performance metrics were generated for this dataset.


\subsection{Benchmarked Models}
Due to having twelve models in our benchmark, for conciseness, we will offer a brief summary of the models' architecture and properties without diagrams or equations. More details on the respective architectures can be found in the original publications. Details on the computational requirements of these models can be found in Table \ref{tab:parameters}.

\subsubsection{CNN-based models}
Convolutional Neural Network (CNN) in semantic segmentation are designed to classify each pixel in an image into a specific category, thus creating a detailed pixel-wise segmentation of the image. By utilizing convolutional layers to extract features at different spatial scales and pooling layers to downsample the features, CNNs can reasonably capture both local and global context information within an image. Additionally, the use of skip connections, such as in U-Net architecture, helps preserve spatial information and improve segmentation accuracy. Overall, CNNs have proven to be highly effective for semantic segmentation tasks in various domains such as medical imaging, autonomous driving, and satellite imagery analysis

\noindent \textbf{FCN (Fully Convolutional Network):} This model modifies CNNs to process full images directly. By substituting fully connected layers with convolutional layers, the model outputs spatial maps suitable for inputs of any size, a significant advancement over traditional CNNs that demand fixed-size inputs (Long et al., 2015). This "fully convolutional" design enables adaptable and size-invariant segmentation. 

\noindent\textbf{FastFCN (Fast Fully Convolutional Network):} This model introduces critical innovations over conventional CNN approaches (Wu et al., 2019). Its main advancement is the Joint Pyramid Upsampling (JPU) module, which efficiently merges multi-scale features, bypassing the extensive pooling and upsampling layers typical of CNNs and FCNs. This allows for high-resolution semantic segmentation with reduced computational demand, facilitating quicker processing speeds than traditional models.

\noindent \textbf{U-Net:} This model leverages a convolutional neural network (CNN) in a symmetric encoder-decoder framework (Ronneberger et al., 2015). It adds skip connections which concatenate feature maps from the encoding path with the decoder's upsampled output, enhancing detail localization. This structure effectively maintains spatial context, addressing the common challenge of detail loss in deeper layers found in traditional CNN segmentation approaches.

\noindent \textbf{MobileNetV3:} This model innovates within CNN architectures, optimizing for mobile device constraints without compromising performance (Howard et al., 2019). By integrating Hardware-Aware Network Architecture Search (NAS) and the NetAdapt algorithm, the algorithm fine-tunes its structure for optimal functionality on mobile CPUs. MobileNetV3 introduces architectural enhancements, such as the Lite Reduced Atrous Spatial Pyramid Pooling (LR-ASPP), to improve semantic segmentation efficiency. Additionally, a streamlined segmentation decoder is incorporated to boost performance in dense pixel prediction tasks, ensuring computational efficiency is maintained.

\noindent\textbf{ANN (Asymmetric Non-local Neural Networks):} This model is a CNN-based framework that introduces two innovations: the Asymmetric Pyramid Non-local Block (APNB) and the Asymmetric Fusion Non-local Block (AFNB) (Zhu et al., 2019). APNB reduces computation and memory usage by applying pyramid sampling to non-local blocks, maintaining performance while addressing the high resource demands of traditional non-local operations. AFNB improves segmentation by fusing multi-level features and addressing long-range dependencies, overcoming typical CNN limitations in capturing these dependencies efficiently.

\noindent\textbf{DeepLabV3:} This model marks a notable development in semantic segmentation, integrating atrous convolutions and Atrous Spatial Pyramid Pooling (ASPP) within a CNN architecture (Chen et al., 2017a). Atrous convolutions are employed to broaden the receptive field, preserving the resolution of feature maps, and enhancing the model's ability to assimilate expansive contextual details without downsampling. The ASPP module leverages atrous convolutions at varied dilation rates to efficiently capture information across multiple scales, ensuring precise segmentation of objects of different sizes.

\noindent\textbf{DeepLabV3+:} This model enhances DeepLabV3 by adding an encoder-decoder structure for better detail and edge precision in semantic segmentation (Chu et al., 2021). It improves on outlining object boundaries and pixel labelling by refining the ASPP module with a decoder to efficiently capture object edges. Depth-wise separable convolution in the ASPP and decoder minimizes computational complexity, ensuring efficient, high-performance segmentation.

\noindent\textbf{Pyramid Scene Parsing Network (PSPNet):} This model employs a CNN-based structure, elevating scene parsing through its Pyramid Pooling Module, which aggregates multi-scale contextual information for superior global comprehension (Zhao et al., 2017). Utilizing features from four distinct pyramid scales, it captures a wide array of global details, essential for the precise parsing and interpretation of complex scenes. This strategic approach overcomes the challenge of fusing global contextual insights, significantly boosting scene parsing accuracy by thoroughly analyzing the interconnected relationships present within images.

\textbf{SegNeXt:} This model is CNN-based and introduces a novel convolutional attention mechanism to enhance computational efficiency (Guo et al., 2022). It leverages convolutional operations for spatial hierarchy management and local feature extraction, key for segmentation, avoiding the computational load of transformers' self-attention. This mechanism efficiently encodes spatial context with specialized convolutions, aiming to balance computational and parameter efficiency with high segmentation accuracy across various datasets.

\subsubsection{Transformer-based Models}
The image Transformer is a relatively novel type of deep learning architecture that leverages the transformer-based architecture, originally designed for natural language processing tasks, for semantic segmentation in images. First introduced by \cite{vit}, by treating the image as a sequence of patches and applying self-attention mechanisms, these image Transformers can capture long-range dependencies and context information effectively. This family of model has shown promising results in semantic segmentation tasks by enabling efficient processing of spatial relationships and capturing global context within the image. The image Transformers have the potential to outperform traditional convolutional neural networks in certain segmentation tasks by offering a different approach to feature extraction and context aggregation in images.

\noindent\textbf{Twins-SVT:} This model, particularly through its variants Twins Pooled Convolutional Pyramid Vision Transformer (Twins-PCPVT) and Twins Scaled Vision Transformer (Twins-SVT), revolutionizes spatial attention in vision transformers with a novel and streamlined design (Chu et al., 2021). This innovation is marked by a simplified yet potent spatial attention mechanism that stands in contrast to the complex and resource-intensive approaches of traditional models. By employing a direct and efficient spatial attention strategy, both Twins-PCPVT and Twins-SVT architectures achieve high computational efficiency through optimized matrix multiplications, ensuring robust model performance without the burden of excessive computational demands.

\noindent\textbf{Segmenting Transformers (SegFormer):} This model is transformer-based and capitalizes on self-attention mechanisms to grasp global dependencies for improved scene understanding (Xie et al., 2021). It innovatively omits positional encoding, avoiding issues related to varying input image resolutions during testing. Additionally, SegFormer integrates a lightweight Multi-Layer Perceptron (MLP) decoder to blend multiscale features from the encoder, efficiently marrying local and global context for accurate segmentation outcomes.

\noindent\textbf{Shifted Window (Swin) Transformer:} This model is transformer-based and features a novel shifted window design that diverges from the fixed-size patches (Liu et al., 2021). This design enables adaptive feature extraction across scales - which is essential for semantic segmentation. It achieves linear computational complexity with image size, improving upon the quadratic complexity of standard transformers. Its hierarchical structure facilitates efficient processing of images at different resolutions, effectively extracting local and global features with enhanced accuracy and scalability.

\begin{table*}[!h]
\small
\centering
\caption{Summary of benchmarked models: computational requirements and weight of training objectives}
\label{tab:parameters}
\noindent
\begin{tabular}{l|c|cc|cccc}
\hline
Model       & Base & FLOPS & Parameters& Cross Entropy & Dice & Focal & Tversky \\ \hline
FCN \cite{Long_2015}        & ResNet101 & 276 G & 66.1 M & 1/2    & 1/2    &        &  \\
FastFCN \cite{Wu_2019}     & ResNet50 & 94.6 G & 53.3 M  & 1/2    & 1/2    &        &  \\
U-Net \cite{Ronneberger_2015}      & -        & 203 G & 29.0 M & 1/2    & 1/2    &        &     \\
DeepLabV3 \cite{Chen_2017}  & ResNet101 & 346 G & 84.7 M & 1/2    & 1/2    &        &   \\
DeepLabV3+ \cite{Chu_2021}  & ResNet101 & 254 G & 60.2 M & 1/2    & 1/2    &        &      \\
ANN \cite{Zhu_2019}        & ResNet101 & 263 G & 62.9 M& 1/3    & 1/3   & 1/3 &    \\
MobileNetV3 \cite{Howard_2019} & Large & 8.80 G    & 3.30 M  & 1/2    & 1/2    &        &       \\
PSPNet \cite{Zhao_2017}      & ResNet101 & 256 G & 65.6 M & 1/2    & 1/2    &        &      \\
SegNeXt \cite{Guo_2022}    & Base & 34.5 G & 27.6 M & 10/11    &               &        & 1/11 \\
Twins \cite{Chu_2021}      & Base & 70.2 G & 59.7 M & 10/11    &               &        & 1/11 \\
Swin \cite{Liu_2021}       & Base & 305 G & 120 M  & 1/3    & 1/3   & 1/3 &   \\
SegFormer \cite{Xie_2021}  & Base & 74.6 G & 82.0 M & 1/2    &               &        & 1/2 \\
\hline
\end{tabular}
\end{table*}

\subsection{Experiment Details}
As the core of our transfer learning strategy for twelve models, each utilizing a backbone pre-trained on the ImageNet 1K dataset. These backbones encompass a range of architectures, ResNet-50, ResNet-101, and Vision ViT. Initially, the models were trained on the SkyScapes dataset as a domain transfer. Upon achieving satisfactory performance on the SkyScapes dataset, the models undergo a subsequent phase of fine-tuning for the Waterloo Urban Scene dataset.

The training was conducted on one NVIDIA RTX 3080 GPU with 10 GB memory. Given the intensive computational requirements, a batch size of 2 was opted for most model training setup. The models were trained around 20 epochs, a duration determined to be sufficient for converging to a stable solution without overfitting, based on the validation performance. To enhance the utility of data, neighboring patches were overlapped by 50\% in both horizontal and vertical dimensions.

The selection of learning rates was informed by established practices in pretrained model configurations, ensuring convergence. Tailored to the sensitivity of individual models to large adjustments, specific learning rates were assigned: ANN and Twins used 0.0001; DeepLabV3, DeepLabV3+, FCN, FastFCN, MobileNetV3, PSPNet, U-Net used 0.01; SegFormer, SegNeXt, and Swin used 0.00006.  A uniform strategy encompassed a warm-up phase, spanning approximately half an epoch, during which learning rates were minimized to stabilize initial model states. Subsequently, a Poly Learning Rate schedule, with a power of 0.9, was employed to systematically reduce learning rates, thereby optimizing training progression.

Both AdamW and Stochastic Gradient Descent (SGD) optimizers were utilized, capitalizing on their respective strengths in managing sparse gradients and momentum. AdamW used favored for Transformer models, namely SegFormer, SegNeXt, Swin, and Twins, while SGD was applied to the remaining models. A diverse suite of loss functions, including cross-entropy loss, Tversky loss, dice loss, and focal loss, were combined as weighted sum for model training. The details are summarized in Table \ref{tab:parameters}.

\section{Results and Discussion}
\subsection{Model Performance and Adaptation}
\subsubsection{SkyScapes Dataset}
This section presents a comprehensive analysis of 12 models' performance on the SkyScapes dataset. The evaluation employs a suite of metrics, including mean Intersection over Union (mIoU), mean Accuracy (mAcc), overall Accuracy (aAcc), mean Recall, mean Precision, and mean F1 score, to facilitate a detailed examination of each model's performance.

Based on Table \ref{tab:skyscapes-lane}, transformer-based models outperform CNN-based models. Within the transformer category, models such as SegFormer and Swin achieve superior results compared to traditional CNN models, highlighting the effectiveness of attention mechanisms over non-attention-based approaches. Notably, SegNeXt, even without employing a transformer structure, achieves the second-best result in terms of mIoU among the 12 models through its unique convolutional attention mechanism. 

Generally, Recall exceeds Precision across the models except for Twins and SegFormer. This suggests their predictions tended to minimize false negatives over false positive to achieve high recall, perhaps over predicting pixels as positives. 

All models demonstrate exceptionally high Accuracy, with each model achieving at least 97.55\% mAcc and some nearing 99.92\%. This phenomenon is attributed to the dominant presence of background pixels, where accurate background prediction significantly influences the Accuracy metric, leading to scale imbalance and diminished result sensitivity. 

In conclusion, examining 12 models on the SkyScapes dataset provides valuable insights into how transformer-based and CNN-based models perform across various metrics. The analysis showed that predicting the background class was generally the easiest task for all models, mainly because it makes up most pixels. However, PSPNet and MobileNetV3 stood out for their lower mIoU scores for the background, suggesting they might incorrectly classify more pixels as belonging to other classes.

\subsubsection{Waterloo Urban Scene Dataset}
This section extends the comparative analysis of model performance from the SkyScapes dataset to the Waterloo Urban Scene dataset by employing identical evaluation metrics, aiming to understand how different models perform across diverse urban imaging domains. 

The results section from Table \ref{tab:waterloo-urban} showcases a notable improvement across all metrics on the Waterloo Urban Scene dataset, with mIoU now ranging between 33.56\% to 76.11\% and F1 scores spanning from 44.34\% to 85.35\%. This marks a significant enhancement in model performance compared to Skyscapes benchmarks, potentially attributed to the advantages of pretraining on the SkyScapes dataset and the unique characteristics of the Waterloo Urban Scene dataset itself.

\begin{table*}[ht]
\footnotesize
\centering
\caption{Benchmark of the state-of-the-art on the SkyScapes-Lane task over all 12 classes (in \%)}
\label{tab:skyscapes-lane}
    \begin{tabular}{l|cccccc}
        \hline
    Method & \textbf{IoU Mean} & \textbf{mAcc} & \textbf{aAcc} & \textbf{mRecall} & \textbf{mPrecision} & \textbf{F1}  \\
        \hline
    FCN & 10.68 & 98.87 & 93.21 & 13.53 & 11.38 & 13.53  \\
    FastFCN & 16.33 & 99.77 & 98.61 & 31.10 & 21.02 & 22.36  \\
    U-Net & 14.98 & 99.24 & 95.43 & 50.45 & 18.97 & 21.80  \\
    DeepLabV3 & 10.24 & 98.79 & 92.72 & 35.32 & 11.12 & 12.69  \\
    DeepLabV3+ & 18.08 & 99.84 & 99.02 & 34.62 & 23.58 & 24.91  \\
    ANN & 20.94 & 99.41 & 96.47 & \textbf{66.00} & 22.92 & 29.88  \\
    MobileNetV3 & 11.74 & 98.91 & 93.47 & 54.55 & 12.45 & 15.25  \\
    PSPNet & 8.68 & 97.55 & 85.29 & 32.24 & 10.09 & 10.65  \\
    SegNeXt & 32.26 & 99.86 & 99.14 & 49.96 & 48.98 & 44.20  \\
    Twins & 30.11 & 99.85 & 99.08 & 45.79 & 51.81 & 41.87  \\
    Swin & 30.51 & 99.84 & 99.02 & 60.03 & 40.27 & 42.97  \\
    SegFormer & \textbf{33.56} & \textbf{99.92} & \textbf{99.50} & 43.85 & \textbf{64.33} & \textbf{44.34}  \\
        \hline
    \end{tabular}
\end{table*}

\begin{table*}[ht]
\footnotesize
\centering
\caption{Benchmark of the state-of-the-art on the Waterloo Urban Scene Dataset over all 15 classes (in \%)}
\label{tab:waterloo-urban}
\begin{tabular}{l|cccccc}
\hline
Method & \textbf{IoU Mean} & \textbf{mAcc} & \textbf{aAcc} & \textbf{mRecall} & \textbf{mPrecision} & \textbf{F1} \\
\hline
FCN & 44.40 & 99.58 & 96.84 & 93.55 & 47.06 & 56.38 \\
FastFCN & 50.48 & 99.76 & 98.16 & 98.54 & 51.46 & 62.07 \\
U-Net & 44.64 & 99.52 & 96.37 & 91.14 & 46.95 & 56.62 \\
DeepLabV3 & 47.55 & 99.63 & 97.22 & 93.97 & 49.44 & 60.31 \\
DeepLabV3+ & 51.03 & 99.65 & 97.34 & 95.34 & 53.20 & 63.74 \\
ANN & 46.50 & 99.67 & 97.54 & 96.30 & 47.91 & 58.23 \\
MobileNetV3 & 34.94 & 99.37 & 95.26 & 89.89 & 37.44 & 46.03 \\
PSPNet & 50.48 & 99.72 & 97.87 & 96.73 & 51.90 & 62.29 \\
SegNeXt & 65.77 & 99.66 & 97.45 & 96.81 & 67.24 & 77.60 \\
Twins & 62.86 & 99.23 & 94.24 & 88.15 & 68.64 & 76.01 \\
Swin & 60.22 & \textbf{99.84} & \textbf{98.78} & \textbf{98.96} & 60.84 & 72.48 \\
SegFormer & \textbf{76.11} & 99.77 & 98.27 & 97.94 & \textbf{77.44} & \textbf{85.35} \\
\hline
\end{tabular}
\end{table*}

In line with the trends observed on the SkyScapes dataset, transformer-based models continue to outperform traditional CNN-based models on the Waterloo Urban Scene dataset. Specifically, SegNeXt, with its convolutional attention mechanism, achieves the second highest mIoU, trailing only behind SegFormer. SegFormer leads in mIoU, mean precision, and F1 score, with impressive scores of 76.11\%, 77.44\%, and 85.34\%, respectively. Swin, on the other hand, excels in mAcc, aAcc, and mean recall, recording the highest values at 99.84\%, 98.74\%, and 98.96\%, respectively. Most models demonstrate exceptionally high recall values, exceeding 90\%, with the exception of MobileNetV3 and Twins, which record slightly lower recalls at 89.89\% and 88.15\%, respectively. This trend suggests that models, in general, tend to predict more pixels outside the actual ground truth area than fewer pixels within it, as evidenced by the lower precision scores compared to recall scores.

Drawing on the insights from the SkyScapes dataset, the analysis of model performance on the Waterloo Urban Scene dataset reveals similar trends. Align with the findings from the SkyScapes dataset, the analysis of models on the Waterloo Urban Scene dataset reveals enhanced performance in the sequence of Road, Traffic Island, Sidewalk, and Vehicle classes which showed larger presence in the imagery. 

This shift in performance suggests a different weighting of class importance within the Waterloo Urban Scene dataset. With playing a more significant role in overall model evaluation metrics, these four classes highlight the importance of dataset diversity in understanding model behavior. These four larger object classes directly attributed to the effectiveness with their alignment on the core objectives of semantic segmentation models tailored for aerial view scene analysis. Conversely, classes characterized by smaller or linear features, like Dash Lines and Long Lines, showcase a opposite trend, with less significant performance enhancements, suggesting the key issue that needs to be overcome in road lane markings segmentation task.

\subsection{Visualization of Results }
\subsubsection{SkyScapes Dataset}
This section provides visual representations of the predictions generated by 12 models on the SkyScapes dataset. Through side-by-side comparisons with input images and ground truth labels. 

Among the visualized predictions, the initial focus is drawn towards PSPNet and MobileNetV3, where misclassification errors are significant, particularly in the background class. This observation aligns with their previously noted lower accuracy. Conversely, despite SegNeXt achieving the second highest mIoU, it is noteworthy that transformer-based models such as SegFormer and Swin exhibit superior visualization results. These transformer-based models demonstrate remarkable fidelity to the ground truth, exhibiting minimal distortion in lane thickness and negligible irrelevant errors. Furthermore, U-Net's performance is notable for its adeptness in delineating boundaries; however, it appears to weak in classifying multiple classes. 

These findings underscore the strengths and weaknesses inherent in each model's architecture and highlight the importance of considering the original design objectives when assessing their performance across diverse tasks. Such insights gained from the visual analysis offer valuable perspectives for further refinement and optimization of semantic segmentation models.



Conversely, all transformer-based models, along with SegNeXt, tend to create lane markings that are finer and closer to the ground truth. Most models are accurate in capturing both long and dashed lines. The detection of zebra zones presents a challenge for several models; however, DeepLabV3+, SegFormer, SegNeXt, Swin, and Twins demonstrate the capability to accurately recognize the zebra zone.

However, some models, particularly MobileNetV3 and U-Net, struggle with a high rate of background misclassification, while DeepLabV3 and FCN also exhibit minor issues with background misclassification. This issue with background misclassification is further validated by Table \ref{tab:skyscapes-lane}, where these models' background mIoU is low.

\subsubsection{Waterloo Urban Scene Dataset}
\begin{figure*}[!t]
\centering
\includegraphics[width=6.8in]{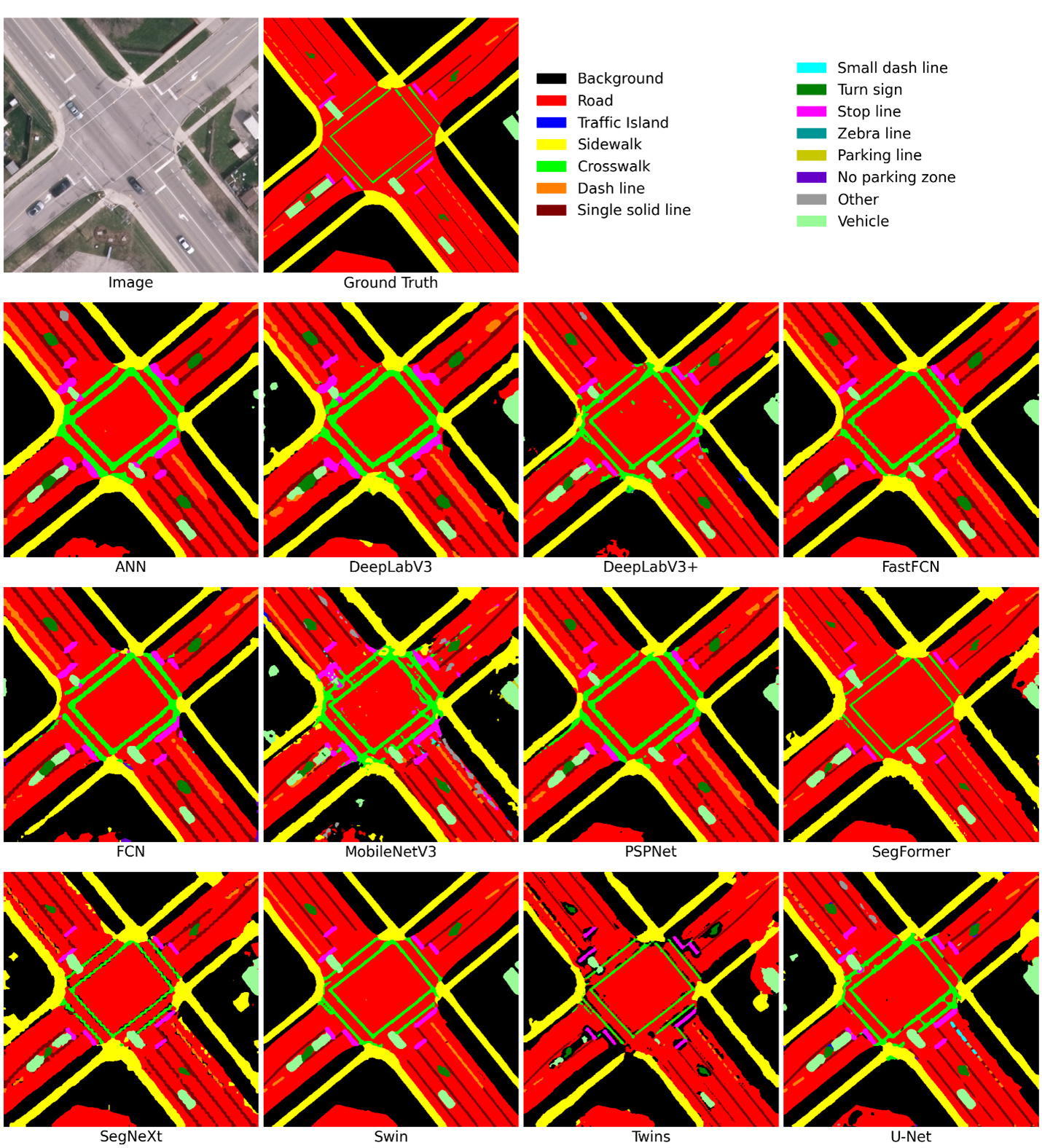}
\caption{A comparative visualization of road lane detection at an intersection by 12 different models on a Waterloo Urban Scene dataset sample. Each model's output is showcased alongside the original image and the ground truth for reference.\label{Picture2}}
\end{figure*}

In the complex intersection scenario shown in Figure \ref{Picture2}, the road lane classifications include features such as stop lines, crosswalks, turn signs, single solid lines, dashed lines, and small dashed lines. The figure also presents classes for vehicles, sidewalks, and roads. Given the scene's complexity, while many models capture the overall appearance, certain elements like turn signs are depicted with excessive thickness, making them difficult to recognize. Swin stands out by accurately classifying most of the scene without significant background misclassifications, although small dashed lines and turn signs are somewhat indistinct. SegFormer also performs well, effectively identifying road lanes despite some confusion in classifying certain background areas as roads. This is deemed acceptable given that these areas share similar color and texture with the road, as verified by the ground truth. MobileNetV3, U-Net, and Twins appear to be the least effective models, showing misclassifications and inconsistencies in scene interpretation, especially when roads are covered by shadows or vegetation.

\subsection{Discussion}
\subsubsection{Dataset and Annotation Quality}

Differences in datasets, as observed between the Waterloo Urban Scene and SkyScapes datasets, can significantly influence model performance. A more balanced distribution, as seen in the Waterloo Urban Scene dataset, facilitates better learning by the model, leading to improved performance. Additionally, higher-resolution datasets like Waterloo Urban Scene may present larger class objects, simplifying the learning task for models designed for aerial-view applications. Understanding dataset characteristics is crucial for optimizing model training and performance evaluation in remote sensing tasks. When examining pixel counts across datasets, it's evident that the Waterloo Urban Scene Dataset encompasses a wider range of classes, and less skewed class distribution contributing to the observed differences.

\begin{figure*}[!t]
\centering

\minipage{0.32\linewidth}
  \includegraphics[width=\linewidth]{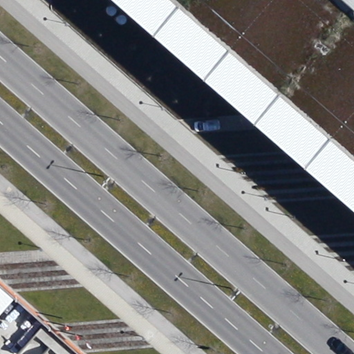}
  \caption{A raw aerial image from the SkyScapes dataset where trees and utility poles obstruct the visibility of road lane markings.}\label{fig3}
\endminipage\hfill
\minipage{0.32\linewidth}
  \includegraphics[width=\linewidth]{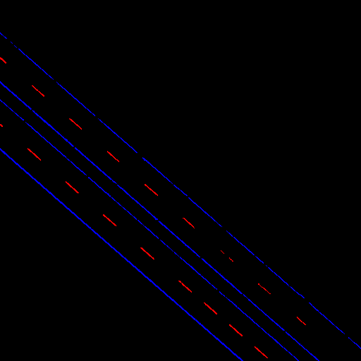}
  \caption{The ground truth image, illustrating that road lane markings are annotated as discontinuous rather than continuous due to obstructions.}\label{fig4}
\endminipage\hfill
\minipage{0.32\linewidth}
  \includegraphics[width=\linewidth]{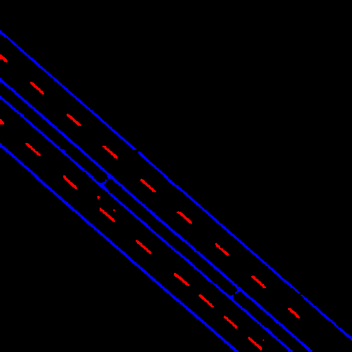}
  \caption{Prediction result from the Swin model on the SkyScapes test dataset, showing relatively continuous road lane markings.}\label{fig5}
\endminipage
\end{figure*}

Aerial images frequently include various obstructions such as trees, vehicles, and shadows, which can obscure critical details. Figure \ref{fig3} illustrates how background features in aerial views, such as road lane markings, often become obscured by trees and utility poles. Consequently, the SkyScapes dataset adopts a questionable annotation practice that only marks road lane markings that are visible in the aerial images, ignoring their actual existence. This method leads to ground truth annotations and training data where road lane markings appear fragmented, as depicted in Figure \ref{fig4}. 


Despite these challenges, some models have demonstrated an ability to overcome such data limitations. Figure \ref{fig5} showcases how the Swin model, for example, effectively discerns the spatial relationships between adjacent road lane marking pixels. This capability allows the model to reconstruct the continuity of road lane markings, thus compensating for the gaps and discontinuities present in the source aerial images. This adaptability highlights the potential of advanced deep learning models to mitigate the effects of noise in aerial imagery, thereby enhancing the reliability of the data derived from these images for various practical applications.

\subsubsection{Model Architecture and Benchmark Results}
Transformer-based models exhibit an advantage in capturing long-range dependencies, which is crucial for understanding complex scenes in remote sensing imagery. Unlike CNNs, which primarily focus on local dependencies, transformers can efficiently learn relationships between distant pixels. However, transformers require intensive training to extract features effectively. The absence of inherent knowledge about pixel distributions and local relations necessitates pre-trained backbones for transformers to achieve robust performance, particularly in smaller datasets with simpler scenes.

Recent research has shown that CNNs can emulate the long-range dependency capturing capability of transformers through the integration of attention mechanisms \cite{Liu_2022}. Models like SegNeXt demonstrate the efficacy of this approach, suggesting that CNNs can rival transformers in certain tasks \cite{Guo_2022}. Additionally, incorporating special pooling layers after CNNs can further enhance performance while reducing computational complexity and runtime.

Both are fundamental metrics in evaluating semantic segmentation performance. As recall increases, precision tends to decrease, and vice versa. For example, in SkyScapes dataset, ANN demonstrates a recall of 66.0\% and a precision of 22.92\%, while SegFormer, a transformer-based model, exhibits a recall of 43.85\% and a precision of 64.33\%. Achieving a balance between these metrics is essential for accurate detection. Strategies to improve recall and precision include refining prediction boundaries to better match ground truth objects and minimizing over-predictions or under-predictions. Utilizing the F1 score, which combines both recall and precision, provides a comprehensive assessment of model performance, particularly in tasks where balancing these metrics is challenging. IoU and F1 score evaluate the overlap between predicted and ground truth regions, considering both shape and location. They provide valuable insights into the accuracy of object detection algorithms. Similar trends can be observed whereby models with high precision and recall also have high mIoU and F1 scores. Accuracy may be skewed by class imbalances, particularly in datasets where certain classes dominate, such as backgrounds in remote sensing imagery. This dominance inflates accuracy scores, potentially masking performance issues in other classes.

\subsubsection{Transfer Learning and Partially Labelled Dataset}

In our experiments, we observed many models were capable of achieving high performance after a simple transfer learning based on full supervision on the initial pretraining dataset. The performance of models on the Waterloo Urban Scene dataset in general surpassed their performance on Skyscape dataset and previous state-of-the-art road lane extraction results. That is to say, given we were able to achieve excellent results on a partially labelled dataset by pretraining on an existing dataset, as such, believe the transfer learning to be a worthwhile step to undertake. We also note that both pretraining and fine-tuning shared a high degree of similarity when comparing to other possible transfer learning schemes such as pretraining on an unlabelled dataset, or a non-remote sensing dataset, then fine-tuning on the Waterloo Urban Scene dataset, and that these other transfer learning schemes can be investigated if no similar fully labeled supervised pretraining datasets exist.

In the absence of dataset labels, it is also possible to consider generative AI, which could possibly be highly beneficial for enhancing the detection and extraction of lane markings. Specifically, diffusion models have the potential to transform the training process for 2D models in lane detection, by synthesizing/adapting training images with different conditions, such as fluctuations in lighting, weather changes, and the presence of dynamic obstacles, as well as be able to transfer and generate annotations for these generated images. 
\section{Conclusion}

This research presents a comprehensive comparative analysis of 12 state-of-the-art CNN- and transformer-based models for road lane extraction using semantic segmentation. The models were evaluated on the SkyScapes Dataset and further fine-tuned on the newly annotated Waterloo Urban Scene Dataset. Transformer-based models, such as Swin and SegFormer, demonstrated superior performance, particularly with common lane markings like solid and dashed lines. Despite challenges posed by noise, such as trees and shadows in aerial imagery, certain models effectively captured the continuity of lane markings by discerning pixel relationships.

The findings offer valuable insights into the strengths and limitations of both CNN and transformer models in the specific context of road lane extraction from aerial imagery. This research not only benchmarks these models but also provides a foundation for future work aimed at refining road lane extraction algorithms and improving datasets for HD map development.

In conclusion, this analysis highlights the capabilities of different semantic segmentation models and underscores the importance of model selection for improving road lane extraction in autonomous vehicle navigation systems. It sets a solid foundation for advancing HD mapping, contributing to more reliable and efficient navigation for autonomous vehicles.
\section*{Acknowledgments}
We thank the members of the Geospatial Intelligence and Mapping (GIM) Lab at the University of Waterloo for their contribution of annotation to the Waterloo Urban Scene Dataset, and Deutsches Zentrum für Luft- und Raumfahrt (DLR) for granting us access to and use of the SkyScapes Dataset.




%

\bibliographystyle{IEEEtran}
\bibliography{References}

\vfill

\end{document}